\documentclass{article}
\usepackage{ijcai26}

\usepackage{times}
\usepackage{soul}
\usepackage{url}
\usepackage[hidelinks]{hyperref}
\usepackage[utf8]{inputenc}
\usepackage[small]{caption}
\usepackage{graphicx}
\usepackage{amsmath, amssymb}
\usepackage{amsthm}
\usepackage{geometry}
\usepackage{booktabs}
\usepackage{algorithm}
\usepackage[switch]{lineno}
\usepackage{listings}
\usepackage{xcolor}
\usepackage{algpseudocode}

\newcommand{\ppwm}[0]{CASSANDRA}
\newcommand{\methodname}[0]{\ppwm}
\newcommand{\methodshort}[0]{CASS}
\newcommand{\bx}{\mathbf{x}}
\newcommand{\bs}{\mathbf{s}}
\usepackage{balance} %

\usepackage{hyperref}
\usepackage{cleveref} %
\usepackage{bbm} %

\title{\methodname: Programmatic and Probabilistic Learning and Inference for Stochastic World Modeling}

\author{
Panagiotis Lymperopoulos$^{1,2}$
\and
Abhiramon Rajasekharan$^{1,3}$\and
Ian Berlot-Attwell$^{1,4,5}$\and
Stéphane Aroca-Ouellette$^1$
\And Kaheer Suleman$^1$\\
\affiliations
$^1$Skyfall AI, 
$^2$Tufts University, 
$^3$University of Texas at Dallas, 
$^4$University of Toronto, 
$^5$Vector Instiute \\
\emails
Corresponding author: ian@skyfall.ai
}

\begin{document}

\maketitle

\begin{abstract}

Building world models is essential for planning in real-world domains such as businesses. Since such domains have rich semantics, we can leverage world knowledge to effectively model complex action effects and causal relationships from limited data. In this work, we propose \methodname, a neurosymbolic world modeling approach that leverages an LLM as a knowledge prior to construct lightweight transition models for planning. \methodname~ integrates two components: (1) LLM-synthesized code to model deterministic features, and (2) LLM-guided structure learning of a probabilistic graphical model to capture causal relationships among stochastic variables. We evaluate \methodname\ in (i) a small-scale coffee-shop simulator and (ii) a complex theme park business simulator, where we demonstrate significant improvements in transition prediction and planning over baselines.
\end{abstract}

\section{Introduction}
\begin{figure*}
    \centering
    \includegraphics[width=0.8\linewidth]{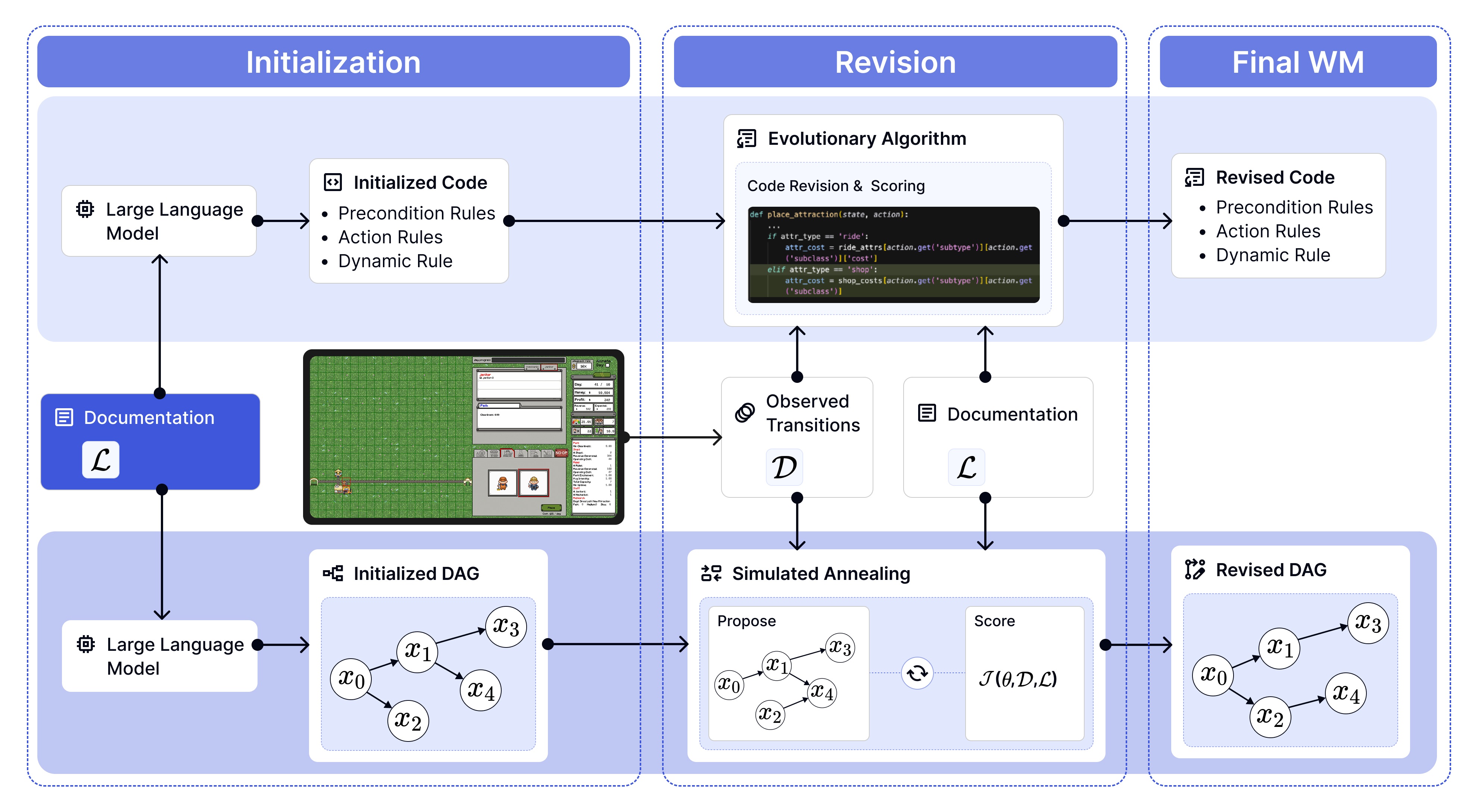}
    \caption{Overview of the \ppwm~ architecture. Deterministic dynamics are modeled by LLM-generated code, which is optimized by an evolutionary algorithm using observed trajectories (top). Stochastic dynamics are modeled by a probabilistic graphical model, designed through simulated annealing structure search with an LLM prior and observed transitions (bottom).}
    \label{fig:framework}
\end{figure*}

\label{sec:introduction}

Effective planning over extended horizons requires an internal model of the environment \cite{richens2025general}. 
However, developing such a world model for real-world environments is challenging, as these domains often contain both deterministic and stochastic dynamics.
Consider an agent for a pizza business deciding whether to purchase a better oven. This action has \emph{deterministic} effects: paying for the new oven and reduction in pizza preparation time. It also has cascading \emph{stochastic} effects: faster preparation may improve customer satisfaction and demand, which in turn affects sales and revenue. An accurate world model must therefore capture both deterministic and stochastic consequences of actions from limited data---a capability that many existing benchmarks overlook \cite{hafner2021benchmarking,chollet2019measure,jansen2024discoveryworld}.

Recent advances in large language models (LLMs) have led to their application in world modeling from limited data, but existing approaches still fall short in domains with mixed dynamics.
Using pre-trained LLMs directly as zero-shot world models \cite{llm-wm-mcts} fails on two fronts: (i) they are ill-suited for the precise symbolic computation (e.g., arithmetic) required to model deterministic variables \cite{bai2025canttransformerslearnmultiplication,llm-text-wm}, and (ii) such models often do not capture domain-specific data distributions \cite{schoenegger2023largelanguagemodelprediction}.
Code world models \cite{worldcoder,codewm-mcts} address the symbolic computation issue by executing LLM-written code, which is more lightweight and sample-efficient. However, existing methods focus on deterministic environments and lack a principled mechanism for fitting models to stochastic real-world data \cite{worldcoder}.

To address these limitations, we introduce \methodname ~(Combining A Symbolic and Stochastic Architecture for Non-deterministic Domains), a dual-stream neurosymbolic framework that integrates a symbolic code component for deterministic dynamics with a probabilistic graphical model for stochastic uncertainty. As shown in Figure~\ref{fig:framework}, our approach leverages a natural language description to initialize both the programmatic model and the graphical structure. Both components are then refined using observed trajectories.

We compare \ppwm~ against baselines including WALL-E \cite{walle}, WorldCoder \cite{worldcoder}, and targeted ablations that isolate the impact of code refinement (deterministic component) and causal structure learning (stochastic component). We evaluate each world model on transition prediction and downstream planning performance when integrated with an agent in two business simulators with non-trivial stochastic dynamics: CoffeeShopSimulator and Mini Amusement Parks (MAPs) \cite{arocaouellette2025miniamusementparksmaps}.

Our main contributions are: (1) \methodname, a novel dual-stream world modeling framework that integrates deterministic and stochastic components initialized from natural language; (2) a hybrid refinement methodology that improves the two components using observed trajectories; and (3) extensive experiments showing that \ppwm~ significantly improves transition prediction and planning over baselines.

\section{Related Work}

Recent work leverages large language models (LLMs) to improve the sample efficiency of learning world models \cite{zhao2023large}. A common approach uses an LLM directly as a world model for next-state prediction, often combined with planning methods such as Monte Carlo Tree Search \cite{llm-wm-mcts}. However, LLM world models struggle with domain-specific transition dynamics requiring arithmetic or symbolic precision \cite{llm-text-wm}, and they are unreliable for verifying action preconditions and effects \cite{llm-wm-precond-effect}. WALL-E \cite{walle} partially addresses this by learning code-based precondition rules, but still relies on LLM state prediction. This is unreliable since untuned LLMs are weak at probabilistic forecasting \cite{schoenegger2023largelanguagemodelprediction}, and fine-tuning remains data-hungry \cite{llm-ft-embodied}. In addition, repeated LLM calls impose high inference costs and latency for planning.

A complementary direction is \emph{code world models}, which use LLM-generated programs to model transition dynamics. These models are lightweight and can be more sample-efficient. WorldCoder \cite{worldcoder} generates a code world model through environment interaction, but assumes deterministic environments and refines a monolithic program. In contrast, \ppwm~ uses targeted code refinement based on the transition prediction error. Other related systems include Visual Predicator \cite{predicator}, which learns action preconditions and effects but focuses on visual domains, and AlphaEvolve \cite{alpha-evolve}, which evolves code via refinement but doesn't address stochastic dynamics. For probabilistic settings, PoE \cite{poe-wm} and OneLife \cite{khan2025lifelearninferringsymbolic} learn probabilistic code world models, but are limited to categorical distributions over code fragments and remain untested in complex stochastic environments. Probabilistic program induction has also been explored \cite{ellis-probab-program}, though existing approaches struggle to fit rich distributions and are mainly demonstrated in simple domains.

Finally, LLMs can contribute causal priors for stochastic modeling. Prior work shows LLMs support causal reasoning \cite{ma2024causal,jiralerspong2024efficient,ban2025llm} and language-guided causal discovery \cite{choi2022lmpriors,requeima2024llm}. To our knowledge, \ppwm~ is the first framework to explicitly leverage LLM-based causal discovery in world modeling.

\section{Methodology}
Our goal is to learn an expressive, hybrid world model of an environment from environment documentation and observed trajectories.

\subsection{Framework Overview}\label{section:framework-overview}
We model the environment as a Semantically-Rich Partially Observable Markov Decision Process (SR-POMDP) defined by
$(\mathcal{O}, \mathcal{S}, \mathcal{A}, \mathcal{P}, \mathcal{R}, \gamma, \mathcal{L})$,
where $\mathcal{O}$ is the observation space, $\mathcal{S}$ the state space, $\mathcal{A}$ the action space, $\mathcal{P}$ the transition function, $\mathcal{R}$ the reward function, $\gamma$ the discount factor, and $\mathcal{L}$ is a text corpus containing semantic descriptions of the observation variables and environment dynamics.

At each time step $t$, the observation is a vector consisting of deterministic and stochastic components, $(\bs_t, \bx_t)$. We express both the ground-truth transition distribution $\mathcal{P}$ and reward function $\mathcal{R}$ in terms of these observables, and include the reward $r_t$ as an additional stochastic observable. We also observe an action $a_{t-1}$ and action validity $\sigma_{t-1}$, which captures whether the action can be executed in the corresponding state. We assume $\sigma_{t-1}$ is deterministically computed. Finally, we learn a world model from a dataset of observed transitions 
$\mathcal{D}=\{(\bs_{t-1},\bx_{t-1},a_{t-1},\sigma_{t-1},\bs_t,\bx_t)_i\}_{i=1}^N$.

By separating deterministic from stochastic variables, we can factorize the transition probability 
as follows:
\begin{multline}
    p(\bx_t, \bs_t| \bs_{t-1}, \bx_{t-1}, a_{t-1}) =\\ p(\bx_t | \bx_{t-1}, a_{t-1}, \bs_{t}, \bs_{t-1}) p(\bs_{t} | \bs_{t-1}, \bx_{t-1}, a_{t-1}).
\label{eq:joint_transition}
\end{multline}
This separation enables the use of specialized algorithms that best model each component. The primary goal of our work is to learn the two parameterized approximations $p_{\theta}(\bx_t | \bx_{t-1}, a_{t-1}, \bs_{t},\bs_{t-1})$, and $p_{\phi}(\bs_t | \bs_{t-1}, \bx_{t-1}, a_{t-1})$ where the latter is defined as the indicator function, $\mathbbm{1}_{\{f_{\phi}(\bs_{t-1}, \bx_{t-1}, a_{t-1})\}}(\bs_t)$ using a deterministic function $f_\phi$.
We optimize $\theta$ and $\phi$ using the environment prior knowledge in $\cal L$, and the observed transitions $\cal D$.

In the following subsection, we describe our strategy for each of the two optimization problems.

\subsection{Modeling Deterministic Dynamics using LLM-Generated Code}
We model direct action effects and environment-driven deterministic changes in the state using code. 
The code $\phi$ can be executed as a function denoted as $f_\phi$ to produce the following: 1) the deterministic observation vector, $s_t$, and 2) action validity boolean, $\sigma_{t-1}\in\{True, False\}$ that indicates whether the action $a_{t-1}$ is valid for the given observation.

\subsubsection{Code Components:}
The deterministic observation vector $s_t$ is obtained by applying two sources of deterministic change to the previous observation. First, the action $a_{t-1}$ induces direct effects on $(s_{t-1}, x_{t-1})$, which we model via \textbf{action functions}, \texttt{a}. Second, the environment induces action-independent deterministic updates (e.g., advancing the time step or paying fixed salaries), which we capture with a \textbf{dynamic function}, \texttt{d}. Since LLM-based world models predict state \emph{differences} more reliably than full next states \cite{llm-text-wm}, we design these functions to output state changes which together transform $s_{t-1}$ into $s_t$.

To model action validity $\sigma_t$, we require action-specific checks that must all pass for an action to be executable. We therefore introduce \textbf{precondition functions}, \texttt{pc}, with one function per action-specific check. Together, the action (\texttt{a}), dynamic (\texttt{d}), and precondition (\texttt{pc}) functions form the deterministic model's code $\phi$.

Breaking down the code into multiple purpose-specific functions allows the LLM to focus on the right functions during revision, enabling \textit{targeted refinement} using a smaller context size.

\subsubsection{Code Optimization:}\label{subsection:code_optimization} We generate the three types of functions that comprise $\phi$ using LLMs (prompt templates are given in the Appendix \ref{appendix:code_prompts}). This code is optimized in two steps: 1) Initializing $\phi$ using $\mathcal{L}$ and 2) Refining $\phi$ using $\mathcal{L}$ and $\mathcal{D}$. These two steps are explained below:

\noindent \textbf{Initializing code:} We use a prompt $P_{init}$ with the game documentation information $\mathcal{L}$ to generate an initial version of code $\phi^0$.

\noindent \textbf{Refine code:} The initial code may be far from perfect, especially in complex domains. Revising the code using trajectory data ($\mathcal{D}$) helps correct errors in the code and fill in the missing gaps \cite{worldcoder,codewm-mcts,alpha-evolve}. We follow an evolutionary algorithm guided by a heuristic scoring function to iteratively improve $\phi$ as explained below.

 Given an observed transition and the code, $\phi'$, we can compute the predicted deterministic observation and action validity: $s'_t, \sigma'_{t-1} = f_{\phi'}(s_{t-1}, x_{t-1}, a_{t-1})$.
Comparing ($s'_t, \sigma'_{t-1}$) with the true values, ($s_t, \sigma_{t-1}$), we identify four types of prediction errors: $E_{exec}$ (Execution/syntax error), $E_{pf}$ (precondition is too restrictive, i.e. $\sigma'_{t-1} = F$ and $\sigma_{t-1} = T$), $E_{ps}$ (preconditions are too permissive, i.e. $\sigma'_{t-1} = T$ and $\sigma_{t-1} = F$), and $E_{od}$ (observation vector is incorrect, i.e. $s'_t \neq s_t$).

Given an error, K possible code refinements are generated. Each refinement is an operation that adds, removes, or replaces a single function in $\phi'$. For each refinement, we compute a score using a heuristic scoring function, $RefinementScore$ ($RS$), and the refinement with the highest score for the current transition is selected to update the code. $RS$ computes the decrease in prediction error for the current transition due to the refinement. However, there's an important caveat. A refinement based on an \textit{incorrect} idea may work well for the current transition, but would not apply to all other transitions in the dataset. Accepting such a refinement may worsen $\phi'$. We mitigate this using a set of validation transitions, V. Only the code refinements that also improve predictions for transitions in V are selected. Additional details of the refinement procedure are given in the Appendix \ref{appendix:refinement_score}.

\subsection{Modeling Stochastic Dynamics with LLM priors}
\label{sec:stochastic}

Our world model's ability to handle uncertainty lies in learning the conditional distribution $p(\bx_t | \bx_{t-1}, a_t, \bs_{t}, \bs_{t-1})$. A key challenge is that individual variables within $\bx_t$ are unlikely to be conditionally independent given the past. For example, in a business simulator, the number of customers and the daily profit are distinct stochastic variables, but they are clearly dependent on each other, even after conditioning on the previous day. Capturing causal variable dependencies is especially important for planning, as failing to do so can lead to unrealistic states during rollouts. We aim to leverage LLMs' knowledge of causal reasoning \cite{ma2024causal} to aid in structuring the dependencies in $\bx_t$.  

To model these complex dependencies, we utilize Bayesian Networks represented by a Directed Acyclic Graph (DAG). This provides a principled trade-off, capturing conditional dependencies between variables while remaining computationally tractable for training and inference: a natural choice for modeling causal relations.

In this section, we describe our method for searching for effective DAG structures. Our goal is to construct a graph $G=(V,E)$ where $V = V_x \cup V_s$ is a set of nodes such that $v\in V_x$ (corresponding to entries in $\bx_t$), $v\in V_s$ (akin for $\bs_t$), and edges $(v_i,v_j)\in E$ form a DAG,  encoding variable dependencies in $\bx_t$. Also, for a given node $v\in V$, $\mathcal{L}[v]$ corresponds to the name and description of the variable in natural language. 
Given a DAG $G=(V,E)$ and a topological ordering of the variables $O$, the conditional distribution factors as follows:
\begin{multline}
    p_{\theta}(\bx_t \mid \bx_{t-1}, a_{t-1}, \bs_t, \bs_{t-1}) =\\ \prod_{v \in O} p_{\theta_v}(\bx_{t,v} \mid \bx_{t, \text{pa}(v)}, \bx_{t-1}, a_{t-1}, \bs_t, \bs_{t-1}),
\end{multline}
where $pa(v)$ indexes the parents of $v$ in the DAG. Each model $p_{\theta_v}$ is represented by a parameterized distribution (e.g., a neural network) and can all be fit in parallel using observed transitions. In the rest of this section, we discuss how to leverage the environment description $\cal L$ to learn an effective structure $E$ that accurately models variable dependencies. 

\subsubsection{\textbf{DAG structure search}}
We search for causal DAG structures for $p_\theta(\bx_t | \bx_{t-1}, a_t, \bs_{t}, \bs_{t-1})$ by a simulated annealing procedure that searches for structures that fit the data while being consistent with the dynamics described in $\mathcal{L}$. First, we sample initial DAGs using LLMs to inform the structure based on the environment description. This ensures we start from graphs that encode meaningful relationships between variables. Then, we conduct a search guided by empirical data and $\cal L$ to refine it. 

\paragraph{Sampling semantically coherent DAGs.} To initialize our search, we sample DAGs for the variables in $\bx_t$ using a two-stage algorithm. First, we prompt an LLM to create a topological ordering of $V$ using $\cal L$. Then, we iteratively prompt to elicit causal variable dependencies consistent with the topological order. The algorithm is explained further in Appendix ~\ref{appendix:stochastic-implementation}.

\paragraph{Searching for DAGs with data and LLM prior.} Searching the super-exponential space of DAGs is computationally intractable. We therefore restrict the search to structures that both fit the data and are plausible given $\mathcal{L}$. To do so, we construct an LLM-based prior $p(E \mid V, \mathcal{L})$ over edge sets $E$ by casting structure plausibility as a question-answering task. Specifically, we form a token sequence $w(\mathcal{L}, V, E)$ that asks whether the causal relationships encoded by $E$ are consistent with the environment description $\mathcal{L}$. We then use the LLM’s probability of an affirmative answer as the prior probability of the structure during search:

\begin{equation}
    p(E|V,\mathcal{L}) = p_{LLM}(\text{"yes"} | w(E,V,\mathcal{L})).
    \label{eq:prior}
\end{equation}
Additional details for the prior computation and prompts are available in Appendices \ref{appendix:stochastic-implementation} and \ref{appendix:stochastic-prompts}, respectively. 
Next, using the dataset $\cal D$ of trajectories, we can perform structure learning by the following optimization problem:

\begin{equation}
    E_{MAP} = \arg\max_{E} p(D|E)p(E|V,\mathcal{L}),
\end{equation}
where $E_{MAP}$ is the maximum aposteriori estimate of the graph structure with our prior. In practice, we pose this optimization problem as a search and use the following objective function:
\begin{equation}
    \begin{aligned}
        J(E;D) = &\frac{1}{|\cal D|} \log p_\theta(D|E) \\
        &- \lambda_1 \frac{|E|}{|V|^2}+ \lambda_2 \log p(E|V,\cal L).\\
        \label{eq:search_obj}
    \end{aligned}
\end{equation}

The first two terms of equation \eqref{eq:search_obj} are the model score, which encourages fitting the data with simpler models as the number of edges is directly related to the number of parameters (similar to Bayesian Information Criterion). The third term is the LLM prior. The hyperparameters $\lambda_1, \lambda_2$ balance the contributions of different terms. $\lambda_1$ can be used to control sparsity, balancing robustness with model flexibility.

To search for structures using the objective in equation \eqref{eq:search_obj}, we use simulated annealing with random perturbations that preserve acyclicity. Using the samples obtained from the DAG sampling step, we search in parallel. We run multiple searches in parallel, initialized from DAGs sampled in the DAG sampling step. This implicitly accounts for LLM uncertainty: confident substructures appear more frequently across samples and are therefore explored more extensively during search. Additional details on our search procedure are available in the Appendix \ref{appendix:stochastic-implementation}.

\paragraph{Fitting to data} Given a structure $E$, we fit the conditional distributions prescribed by the graph structure using neural networks for the approximation. It is important to fit the distribution of each variable, rather than attempt to predict the mean, as sampling is needed for planning. Therefore, in our experiments, we use quantile regression for continuous variables and train with pinball loss to flexibly model the distribution of values without further assumptions. However, this can be modified if additional domain knowledge is available. We use cross-entropy loss for categorical variables. Additional implementation details are available in the Appendix \ref{appendix:stochastic-implementation}.

\section{Experiments}

In this section, we present experiments on two business simulator environments. We use CoffeeShopSimulator\footnote{Source code will be made available upon publication.} to illustrate our structure learning algorithm and analyze how the learned causal structure impacts planning. We evaluate overall performance on Mini Amusement Parks (MAPs) \cite{arocaouellette2025miniamusementparksmaps} by comparing \ppwm~ against relevant baselines. In both environments, the agent’s current budget is the reward $r_t$, modeled as part of the stochastic state.

To highlight the importance of modeling the causal structure, we include three ablations of \ppwm’s stochastic component: (i) \methodshort-Ind predicts stochastic variables independently, (ii) \methodshort-L replaces the learned DAG with a randomly ordered linear DAG, and (iii) \methodshort-R optimizes a randomly initialized DAG without an LLM prior. The world models are compared as environment simulators by an MCTS planner using random action proposal with simple heuristics. We use a similar search time budget ($\approx 1$ hour per action) for comparison. We evaluate using 36 runs of each agent (see Appendix~\ref{appendix:planners} for additional information). We also compare the deterministic component against an offline implementation of WorldCoder \cite{worldcoder}, and the full system against WALL-E \cite{walle}, adapted to MAPs. We use GPT-4.1-mini for all experiments. Additional implementation details are provided in Appendix~\ref{appendix:baselines}.

\begin{table}[h!]
\small
\centering

\begin{tabular}{lcc}
\toprule
\textbf{Agent} & \textbf{Final Budget}  \\
\midrule
MPC+\methodshort-L & $987.5 \pm 697.5$\\
MPC+\methodshort-Ind & $4853.3 \pm 3689.0$\\
MPC+\methodshort-R & 1669.55 $\pm$ 1475 \\
MPC+\ppwm & $\mathbf{9951.0 \pm 820.0}$\\
\bottomrule
\end{tabular}
\caption{ Budget at the end of 50 days in CoffeeShopSim with different stochastic models.}
\label{tab:coffee-money}
\end{table}

\begin{figure}
    \centering
    \includegraphics[width=0.9\linewidth]{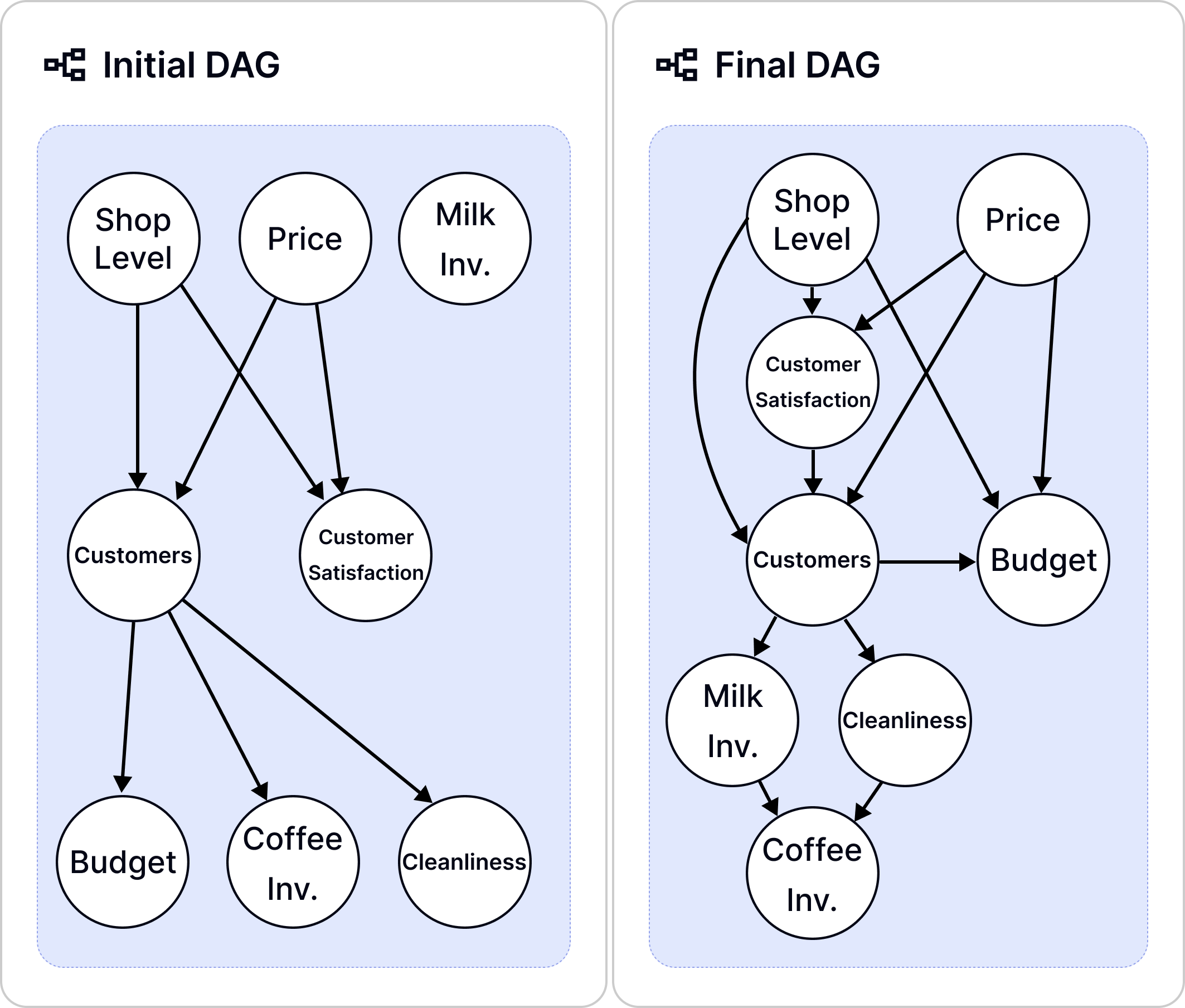}
    \caption{Initial and final DAGs for CoffeeShopSim, modeling dependencies of state variables.}
    \label{fig:coffee-model}
\end{figure}

\begin{table*}[t]
\small
\centering

\begin{tabular}{lccccc}
\toprule
\textbf{Method} & \textbf{10 Days} & \textbf{20 Days} & \textbf{30 Days} & \textbf{40 Days} & \textbf{50 Days} \\
\midrule
MPC+WALL-E-MAPs & 26\% $\pm$ 13\% & 20\% $\pm$ 12\% & 17\% $\pm$ 11\% & 13\% $\pm$ 10\% & 8\% $\pm$ 7\% \\
MCTS+WALL-E-MAPs & 66\% $\pm$ 5\% & 52\% $\pm$ 15\% & 40\% $\pm$ 15\% & 29\% $\pm$ 14\% & 27\% $\pm$ 14\% \\
\hline
MCTS+\methodshort-Ind & 90\% $\pm$ 9\% & 54\% $\pm$ 15\% & 29\% $\pm$ 14\% & 29\% $\pm$ 14\% & 29\% $\pm$ 14\% \\
MCTS+\methodshort-L & 54\% $\pm$ 15\% & 54\% $\pm$ 15\% & 47\% $\pm$ 16\% & 47\% $\pm$ 16\% & 47\% $\pm$ 16\% \\

MCTS+\ppwm & $95\%\pm 5\%$ & $\mathbf{95\%\pm 5\%}$ & $\mathbf{95\%\pm 5\%}$ & $\mathbf{95\%\pm 5\%}$ & $\mathbf{95\%\pm 5\%}$ \\
\bottomrule
\end{tabular}
\caption{Comparison of 95\% confidence interval of survival rates in MAPs over 50 days. Higher values are better.}
\label{tab:bankruptcy_comparison}
\end{table*}

\subsection{Experiment 1: CoffeeShopSimulator}
CoffeeShopSimulator challenges an agent to manage inventory, set prices, and maximize profit over a 50-day horizon. Stochasticity arises primarily from customer behavior and sales, which depend on the store's daily state. This environment provides a controlled setting to evaluate our method's ability to discover a DAG to model these dynamics for effective planning. The experimental setup is explained in more detail in Appendix ~\ref{appendix:coffee-shop-setup}.

\paragraph{Results} Figure~\ref{fig:coffee-model} compares the initial graph with the final optimized structure. The initial graph captures high-level relations but omits key dependencies (e.g., Milk inventory is disconnected). After optimization, the search uncovers a latent relation not explicitly stated in the description: customer satisfaction directly increases same-day customer count. While the learned structure is not perfect (it introduces a spurious edge between coffee and milk inventory), it fits the data well and remains consistent with $\mathcal{L}$. The final DAG and refined code together predict state transitions in \ppwm~, making it more \emph{interpretable} than LLM-call-based world models, such as WALL-E\cite{walle}.

Table \ref{tab:coffee-money} shows the performance of Model Predictive Control (MPC) agents using different stochastic models. Agents using ablated models with a linear or random DAG (\methodshort-L and \methodshort-R) perform poorly by fitting brittle correlations that hinder planning. The agent predicting variables independently (\methodshort-Ind) fares better as predictions are likely more robust, but exhibits high variance due to occasional bankruptcy. The independence assumption here makes the model miss the crucial effects of customer satisfaction on guest count and cleanliness, leading to poor planning. 

\subsection{Experiment 2: Mini Amusement Park (MAPs)}
In the Mini Amusement Park (MAPs) environment \cite{arocaouellette2025miniamusementparksmaps}, an agent starts with a limited budget and must manage the park over a long time horizon by building rides, operating shops, and hiring staff. The primary challenge lies in modeling guest behavior, which is inherently stochastic and dependent on a complex interaction of hidden variables. The state observation of MAPs is a complex JSON object of park information, including park statistics, ride, shop, and staff attributes. While many low-level variables in the JSON are stochastic, we focus our stochastic model on 6 high-impact dependent variables related to planning and use the rest as independent variables to keep the search space tractable. See additional details of the experimental setup in Appendix ~\ref{appendix:map-setup}.

\begin{table}[H]
\small
\centering
\begin{tabular}{lcc}
\toprule
Model &$\sigma_{acc}\uparrow$& $\sigma_{F1}\uparrow$\\ 
\midrule
WALL-E-MAPs & 0.878 & 0.891\\
OfflineWorldCoder &0.905&0.913\\
\methodshort-Init &0.859& 0.872\\
\ppwm &0.945 & 0.948\\
\bottomrule
\end{tabular}
\caption{Comparing \ppwm~against baselines on action success prediction accuracy and F1 score. Higher values are better.}
\label{tab:symbolic-ablation}
\end{table}

\subsubsection{Results}
Table~\ref{tab:bankruptcy_comparison} reports MAPs agent performance measured by survival rate over 50 days. WALL-E-MAPs goes bankrupt in nearly all runs, indicating that it fails to capture the environment’s stochastic dynamics. \methodshort-Ind performs substantially better: its symbolic component accurately models deterministic dynamics, and independent prediction captures some stochasticity. However, most runs still eventually go bankrupt, since ignoring the joint distribution over state variables limits performance under MAPs’ tight margins (e.g., hiring janitors at the right time to maintain cleanliness and revenue). \methodshort-L performs slightly better by capturing limited dependencies. In contrast, \ppwm~ explicitly models causal structure and best captures MAPs dynamics, enabling the agent to survive all 50 days.

Action validity prediction is a crucial component of modeling transition dynamics. Table~\ref{tab:symbolic-ablation} evaluates action success prediction of \ppwm~ against the baselines, including its ablation \methodshort-Init (which omits the code refinement step). \methodshort-init performs substantially worse than \ppwm~, highlighting the importance of the code refinement step. We attribute \ppwm's gains to its targeted code refinement procedure, which more effectively learns action preconditions than WALL-E’s rule learning and WorldCoder’s refinement of a monolithic code block. Additional results for transition modeling are in the Appendix ~\ref{appendix:additional-obs-results}.

\section{Conclusion}

In this work, we introduced \ppwm, a neuro-symbolic framework that leverages an LLM as a knowledge prior to learn programmatic and probabilistic world models. Experiments show that our code refinement learns accurate deterministic dynamics, while LLM-guided structure learning recovers causal dependencies that improve planning in both a coffee shop environment and a complex theme park business simulator.

Despite these promising results, our approach has limitations. First, \ppwm~ assumes a known decomposition of state variables into deterministic and stochastic components; while this may not hold universally, it is natural in many realistic domains. Second, although \ppwm~ avoids LLM calls during planning, it still relies on LLMs for code generation/refinement and structural priors, making performance sensitive to the model’s knowledge and the quality of the environment description. Future work for \ppwm~ can include automatically inferring the deterministic--stochastic decomposition from priors and observed data, improving sample efficiency through active learning, and online refinement to support continual deployment in evolving domains.

\clearpage

\bibliographystyle{named}
\bibliography{ijcai26}

\clearpage

\appendix
\section{Baseline Implementation}\label{appendix:baselines}

\subsection{WALL-E}\label{appendix:baselines:walle}
To adapt WALL-E \cite{walle} for the MAPs environment we modify all the original prompts to incorporate our new state and action specifications. Both the state and action specifications are represented using JSON objects. We further modify the prompt for predicting the next state given the action to incorporate and provide examples of state transitions. For few-shot examples, we associate transitions from the training data with an embedding of the action string, then at inference time we retrieve the top K (5) transitions based on the current action. For rule generation, we follow \cite{walle}; however, unlike in WALL-E, we do not use different outputs for rules predicting valid and invalid actions. All rules will predict True if the action is valid and False if the action is invalid. For learning the rules, we use randomly sampled trajectories with an invalid action error rate of 0.2, however since random trajectories often result in short trajectories due to bankruptcy, we follow a three-step process.  First, we sample a set of trajectories by searching the space for high rewards via Monte-Carlo Tree Search (MCTS) equipped with the ground-truth MAPs environment for rollouts. Second we sample a trajectory and select a random timestep. Finally we rollout using an agent that selects random actions, producing an invalid action with probability 0.2. We believe this would result in something akin to having a small set of suboptimal trajectories.  For MPC game-playing experiments we combine the WALL-E world model with the same rule-based heuristic agent used with CASSANDRA. For each step, we first sample multiple distinct initial actions, then we rollout for k (4) steps. We compute the rewards at the end of each rollout and take the action with the highest-scoring rollout. %

\subsection{WorldCoder}

WorldCoder \cite{worldcoder} is an online approach to learning a deterministic code world model. As our work presents an offline approach to learning a stochastic world model, it was necessary to adapt WorldCoder into a variant we refer to as OfflineWorldCoder. 

The original WorldCoder assumes access to the ground-truth environment, as well as a set of objectives within said environment that it must learn to solve. WorldCoder learns in an online manner, maintaining a replay buffer of transitions observed in the ground-truth environment. It infinitely loops over objectives in the same environment; for each objective it executes a rollout in the environment. In this rollout, actions are selected randomly with probability $\epsilon$. Otherwise, the learned world model and reward functions are used to create a plan satisfying the goal, and the next step of this plan is executed. For most of their environments they use BFS as their planner; for AlfWorld they use MCTS with a manually engineered heuristic for guiding search. If at any time the planner fails to find a plan satisfying the goal, or if the world model becomes inconsistent with a transition in the replay buffer, then code revision is triggered. Revision is via REx code generation \cite{rex}, conditioned on the internal LLM with the existing code. In the case of inconsistency with the replay buffer, a single failing transition is sampled, and the learned transition function and reward function are revised wholly independently. In the case of the planner being unabled to find a solution, they are revised together.

Initially, the system collects purely random data until the replay buffer is sufficiently large, at which point REx is used twice -- once to create the transition function, and again to create the reward function.

First and foremost: as WorldCoder is designed for deterministic environments we only learned a code world model for the deterministic variables of the environment. I.e., we used WorldCoder to optimize the program $\phi$ for the same objective used by our code learning system for deterministic components:

\begin{equation}
    \max_{\phi} \sup_{(\bs_{t-1}, \bx_{t-1}, a_{t-1}, \bs_t, \bx_t) \in \mathcal{D}} \log p_\phi(\bs_t | \bs_{t-1},\bx_{t-1},a_{t-1}),
\end{equation}

As the reward function in the MAPs environment (i.e., profit) is non-deterministic, we did not use WorldCoder for reward learning. This lack of a reward function, combined with the stochastic MAPs environment complicating search, the inability of the WorldCoder model to produce rollouts as it cannot predict stochastic variables, and the nature of our task being optimization rather than satisfaction (thus requiring a conversion to a planning task), means it would require significant engineering and scientific effort to adapt the online training loop. Given that our system is offline and unlike WorldCoder does not assume access to the ground truth environment, we decided that the best course of action was to remove the planning loop and make WorldCoder offline.

In OfflineWorldCoder, the replay buffer is initialized with the full training set, and REx is run. In WorldCoder, REx normally runs until code is generated that fully explains the replay buffer. However, we found that gpt-4.1-mini was unable to learn the function predicting aggregate ride excitement. Given that aggregate ride excitment penalizes duplicate attractions via a geometric decay, and the exact formula is not in the MAPs documentation, it is unsurprising that abducing the rule from a single failed example is virtually impossible. For this reason we instead run REx until convergence, considering REx to be converged if performance does not improve after 10 iterations. Running our experient, REx terminated after the 33rd iteration, failing to improve after iteration 22 (which explained 86.7\% of the training transitions).

\begin{table*}
\small
\centering
\begin{tabular}{lcccccc}
\toprule
Model & $Cat_{acc}^D\uparrow$ & $Num_{rmse}^D\downarrow$ & $Inv^D\downarrow$ & $Num_{rmse}^S\downarrow$ & $Inv^S\downarrow$&$Time(s)\downarrow$\\ 
\midrule
OfflineWorldCoder \cite{worldcoder} &  1.000 & 0.483 & 0.001  & - & - & 3.40\\
WALL-E-MAPs \cite{walle} &   0.944 & 0.084 & 0.042  &  0.515 & 0.157 & 222.30\\
\hline
\methodshort-Ind &  1.000 & 0.176 & 0.012 & 0.254 & 0.122& 8.44\\
\methodshort-L &  1.000 & 0.197 & 0.017 & 0.249 & 0.119&9.59\\
\ppwm & 1.000 & 0.176 & 0.012 & 0.248 & 0.117&10.67\\
\bottomrule
\end{tabular}
\caption{Comparison of world models' transition predictions. $Cat_{acc}$ denotes prediction accuracy for categorical variables. $Num_{rmse}$ denotes the average RMSE of numerical variables. $Inv$ is the average invalid value rate. $Time$ is the average prediction time for one transition. Superscript '$D$' or '$S$' indicates deterministic or stochastic component of the observation, respectively. $\uparrow$ indicates higher number is better and $\downarrow$ indicates lower number is better. (Since OfflineWorldCoder is purely deterministic, we omit stochastic results and enter '-' instead)}
\label{table:wm-eval-obs}
\end{table*}

\section{Additional Experimental results}\label{appendix:additional-results}

\subsection{Experiment 1 Additional Results.}\label{app:additional_coffee}
\small
\begin{table*}[ht]
\centering

\begin{tabular}{l|cccccc}
\hline
\textbf{Method} & \textbf{Money} & \textbf{Coffee} & \textbf{Milk} & \textbf{Customers} & \textbf{Satisfaction} & \textbf{Cleanliness} \\
\hline
\ppwm & $128.8 \pm 2.5$ & $2.5 \pm 0.0$ & $12.6 \pm 0.0$ & $32.6 \pm 0.3$ & $3.1 \pm 0.1$ & $43.5 \pm 0.0$ \\
\ppwm-Ind & $126.1 \pm 3.5$ & $2.5 \pm 0.0$ & $12.5 \pm 0.0$ & $37.0 \pm 0.5$ & $3.2 \pm 0.0$ & $44.3 \pm 0.1$ \\
\ppwm-Lin & $133.2 \pm 1.3$ & $2.8 \pm 0.0$ & $13.1 \pm 0.0$ & $30.3 \pm 0.1$ & $3.0 \pm 0.0$ & $44.0 \pm 0.1$ \\
\ppwm-R & $127.3 \pm 3.1$ & $2.6 \pm 0.0$ & $13.2 \pm 0.0$ & $31.3 \pm 0.2$ & $2.9 \pm 0.0$ & $44.1 \pm 0.1$ \\
\hline
\end{tabular}
\caption{RMSE of predictions in CoffeeShopSimulator}
\label{tab:coffe-shop-accs}
\end{table*}

Table \ref{tab:coffe-shop-accs} shows prediction RMSE of \ppwm~ and ablations in CoffeeShopSimulator. Prediction accuracy is statistically the same across methods. This is because the environment changes relatively smoothly and changes are correlated across timesteps. However, as indicated in the main paper, average prediction accuracy metrics are not sufficient for planning performance, as a model can achieve high accuracy without capturing causal relations between variables. This leads to poor task performance when the world model is deployed for planning. 

\subsection{Modeling Transition Dynamics Additional Results}\label{appendix:additional-obs-results}
To compare the world models based on their transition dynamics, we evaluate the observation vector ($s_t$, $x_t$) predicted by each world model against the true values from 1800 observed transitions drawn from 36 test trajectories of the MAPs environment \cite{arocaouellette2025miniamusementparksmaps}.

To measure observation prediction errors, we use metrics depending on the variable type and error. A prediction can be \textit{invalid} when a variable has a value that is out of expected bounds. We measure an invalid value rate for each variable across all transitions in the test set. We report the average invalid value rate of a set of variables, $Inv$, by averaging the invalid value rates of variables in the set. An observation can have different types of variables (categorical and numerical). Hence, we include average accuracy measurements ($Cat_{acc}$) for categorical variables and average Root Mean Squared Error (RMSE) values ($Num_{rmse}$) for numerical variables (which are first normalized using Min-Max Scaling). Table \ref{table:wm-eval-obs} reports the results for both observation and action validity predictions. The results show that \ppwm~ can model both deterministic and stochastic observation variables effectively in a fraction of the time required by WALL-E-MAPs.

\section{Experimental Setup}

\subsection{System Configuration}
Since the evaluation can be computationally expensive, especially for MCTS rollouts involving an LLM-world model (Wall-E-MAPs), we use AWS Graviton2 instance with 32 vCPUs (16 cores).

\subsection{CoffeeShopSimulation Setup}\label{appendix:coffee-shop-setup}
We first generate a dataset of 100 trajectories using a simple heuristic agent, with 90 used for training. The simulator state contains 8 variables, 6 of which are dependent and modeled by \ppwm. We run 50 search steps from 5 initial samples to find the best-scoring DAG structure. To focus on the stochastic component, all planning evaluations use the ground truth simulator for deterministic effects. 

\subsection{Mini Amusement Park Setup}\label{appendix:map-setup}
We collect 84 high-quality training trajectories from 7 distinct park layouts using an MCTS agent with ground-truth simulator access. All evaluations are performed on 3 held-out layouts.  We construct planning agents by running MCTS using the \ppwm~ model and ablations. Similar to the original WALL-E paper, we embed WALL-E-MAPs in a planning agent using MPC for planning with LLM-based action selection. We also include an MCTS version that uses the WALL-E-MAPs. All agents use a random action proposal with simple heuristics. We use a similar search time budget for all agents ($\approx 1$ hour per action) for comparison. We evaluate using 36 runs of each agent. Additional details are available in the Appendix
~\ref{appendix:planners}.

\subsection{Planning Agents}\label{appendix:planners}

To convert our world models into agents we use Monte Carlo Tree Search (MCTS). To cope with the stochasticity in the environment, we use a MPC approach: using MCTS to produce an optimized plan up to a finite horizon $H$, and re-planning after each action. We also modified our states to instead represent sequences of actions (i.e., a node in the search tree represents the distribution of possible states after a given sequence of actions, rather than a specific state -- to accurately estimate this we perform all rollouts by sampling the world model from the root node to the horizon). We optimize for predicted money and do not use a value function -- i.e., we solely optimize over the timesteps of the horizon. Experiments with MAPs suggest that a horizon of 2 actions is adequate under an oracle world model, but a horizon of a single action fails to capture the benefits of staff hiring.

Due to the high branching factor in MAPs we sample a subset of the possible actions at each timestep and apply basic heuristics to bias the sampled actions towards useful choices. Specifically: 

\begin{itemize}
    \item We enforce that the wait command is always present exactly once in the set of proposed actions
    \item We sample at most one price change action
    \item We do not sample staff movement or guest survey actions
    \item The number of attraction placement and attraction sale actions is not limited
    \item We do not sample invalid actions (e.g., invalid price, positions away from paths, etc...)
    \item We force all ride and shop construction to set prices to their maximum
    \item We bias the sampling of build commands to move the composition of the park towards a {drink shop:food shop:specialty shop ratio} of  {4:3:2} -- this ratio was chosen after playing several games in the MAPs environment 
    \item We constrain $(x,y)$ coordinates to only be sampled from the 5 valid locations nearest the entrance
    \item We de-duplicate actions that are identical up to choice of $(x,y)$ coordinates. Consequently there are at most 2 place staff actions and at most 2 fire staff actions, each for a janitor or mechanic respectively.
\end{itemize}

MCTS hyperparameters for CASSANDRA variants were 90 iterations (i.e., 90 select, expand, simulate, backpropagate rounds), 4 rollouts per frontier node (i.e., simulation was performed with 4 rollouts executed in parallel), 100 sampled actions per node sampled from the heuristic before deduplication, and a horizon of 3 actions. %

For WALL-E there were difficulties due to the slow speed and high-cost of the world model, particularly as inference time increased throughout the trajectory as the state became larger. For this reason is was necesary to increase the time budget per trajectory from 1 hours to 2 hours. Despite this, for it to be possible to run the system under this time budget, it was still necessary to reduce the horizon to 2 actions and the number of iterations to 2. Under these settings the total cost of our MCTS+WALL-E experiments still exceeded \$100USD, with a time to complete the trajectory of 117 minutes. Note that the number of rollouts (4) was kept unchanged, and did not impact runtime as rollouts were performed in parallel. %

As the original WALL-E paper uses MPC over independent rollouts instead of MCTS (possibly due to the slow inference times and high costs we encountered), we also implemented a non-MCTS MPC for WALL-E. The details are outlined in Appendix~\ref{appendix:baselines:walle}.

\section{Code Optimization}
We use gpt 4.1 mini model for both code initialization and refinement to match the baselines (Wall-E-MAPs and OfflineWorldCoder). The prompts used for code initialization and refinement are given in \ref{appendix:code_prompts}.
\begin{algorithm}[H]
\small
\caption{Code Refinement using Observation Transitions}
\label{alg:code_refinement}
\begin{algorithmic}[1]
\Procedure{RefineCode}{$\mathcal{D}$, $\phi'$}
    \For{each transition $(s_{t-1}, x_{t-1}, a_{t-1}, \sigma_{t-1}, s_t, x_t) \in \mathcal{D}$}
        \State $s'_t, \sigma'_{t-1} \gets f_{\phi'}(s_{t-1}, x_{t-1}, a_{t-1})$
        \Comment{Current code predictions}
        \If{$s'_t = s_t \land \sigma'_{t-1} = \sigma_{t-1}$}
            \State \textbf{pass}
        \Else
            \State $\epsilon \gets ErrorType(s'_t, \sigma'_{t-1}, s_t, \sigma_{t-1}), \epsilon \in E$
            \State $P_{refine} \gets T_{refine}(s'_t, \sigma'_{t-1}, s_t, \sigma_{t-1}, \epsilon, \mathcal{L})$
            \State $r^1_t, \dots, r^K_t \gets LLM(P_{refine})$
            \Comment{Generating refinements}
            \State $r^\star_t \gets \underset{RS(s'_t, \sigma'_{t-1}, s_t, \sigma_{t-1}, \epsilon, r^j_t)}{\arg\max} r^j_t$\text{ : } $VS(r^j_t, \phi', \epsilon, V) > 0$
            \State $\phi'' \gets ApplyRefinement(r^\star_t, \phi')$
            \State $\phi' \gets \phi''$
            \Comment{Updating code with best refinement}
        \EndIf
    \EndFor
    \State \Return $\phi'$
\EndProcedure
\end{algorithmic}
\end{algorithm}

\paragraph{Code Initialization}: No observation data is used for code initialization; we only provide game documentation and information needed for generating action, precondition, and dynamic functions. The average token usage for code initialization (for 5 attempts) is \textit{248853} input tokens and  \textit{21790} output tokens. 

\paragraph{Code Refinement}: We select 300 training transitions from 84 training trajectories such that they have a ratio of 1:1 successful ($\sigma=True$) and failing ($\sigma=False$) transitions. We notice that exposing the refinement process to enough failing transitions is effective in learning strong preconditions. These transitions are refined as mentioned in Algorithm \ref{alg:code_refinement}.  Each refinement loop takes an average of 10.9 seconds and has a token usage of \textit{5118} input tokens and \textit{540} output tokens.

\begin{algorithm}[H]
\small
\caption{Code Refinement Heuristic Scoring Function}
\label{alg:refinement_score}
\begin{algorithmic}[1]
\Procedure{PERS}{$\sigma_{v-1}, \sigma_{v-1}', \hat{\sigma_{v-1}}$}
    \State \Return $1_{\hat{\sigma_{v-1}} = \sigma_{v-1}} - 1_{\sigma_{v-1}' = \sigma_{v-1}}$
\EndProcedure
\vspace{0.2cm}
\Procedure{ODRS}{$s_{v-1}, s_v, s_v'$}
    \State $d_1, \cdots, d_L \gets DeepDiff(s_{v-1}, s_v)$\\
    \Comment{Assuming values that are lists or dictionaries are treated as on-numerical entries}
    \State $val^0_1, \cdots, val^0_L \gets GetValues([d_1, \cdots d_L], s_{v-1})$
    \State $val_1, \cdots, val_L \gets GetValues([d_1, \cdots d_L], s_{v})$
    \State $val_1', \cdots, val_L' \gets GetValues([d_1, \cdots d_L], s_{v}')$
    \State $score \gets 0$
    \State $total \gets 0$
    \For{$i \gets 0 to L-1$}
        \If{$DiffType(d_i) \neq values\_changed$}
            \State $score \gets score - 1$
            \State $total \gets total + 1$
        \Else
            \If{$IsNumerical(val_i)$}
                \If{$abs(val_i-val_i') <= abs(val_i-val_i^0)$}
                    \State $score \gets score + 1$
                    \State $total \gets total + 1$
                \Else
                    \State $score \gets score - 1$
                    \State $total \gets total + 1$
                \EndIf
            \Else
                \If{$val_i = val_i'$}
                    \State $score \gets score + 1$
                    \State $total \gets total + 1$
                \Else
                    \State $score \gets score - 1$
                    \State $total \gets total + 1$
                \EndIf
            \EndIf
        \EndIf
    \EndFor
    \State \Return $score / total$
\EndProcedure

\vspace{0.2cm}
\Procedure{RefinementScore}{$r_t^j$, $\phi'$, V}
    \State $funcType \gets FunctionType(r_t^j)$
    \For{$i \gets 0$ to N-1}
        \State $s_{v-1}, x_{v-1}, \sigma_{v-1}, s_v, x_v \gets V[i]$
        \State $s_v', \sigma_{v-1}' \gets f_{\phi'}(s_{v-1},x_{v-1}, \sigma_{v-1})$
        \State $\hat{s_v}, \hat{\sigma_{v-1}} \gets f_{\phi_{r_t^j}'}(s_{v-1},x_{v-1}, \sigma_{v-1})$
        \State $score \gets 0$
        \If{$funcType = PreconditionFunction$}
            \State $score = score + PERS(\sigma_{v-1}, \sigma_{v-1}', \hat{\sigma_{v-1}})$
        \ElsIf{$funcType = ActionFunction \lor funcType = DynamicFunction$}
            \State $score = score + (ODRS(s_v, x_v,\hat{s_v})-ODRS(s_v, x_v,s_v'))$
        \EndIf
    \EndFor
    \State \Return $score / N$
            
\EndProcedure
\end{algorithmic}
\end{algorithm}

\subsection{Code Refinement Implementation}\label{appendix:refinement_score}
The algorithm we use for code refinement is presented in Algorithm ~\ref{alg:code_refinement}. Given the type of error $\epsilon$, a prompt $P_{refine}$ is constructed using template $T_{refine}$ to generate K possible refinements, $r^1_t, \cdots r^K_t$. Each refinement is an operation that adds, removes, or replaces a single function in $\phi'$. Given the candidate refinements, we compute a heuristic scoring function (called $RefinementScore$ ($RS$)) that computes the decrease in prediction error of each refinement on a given transition, compared to the original set of functions. The refinement with the highest score for the current transition is selected to update the code.

The refinement process uses a $RefinementScore$ function as a heuristic. It is of two types, depending on the type of function $r_t^j$ that is being improved: 1) Precondition Error Reduction Score ($PERS$) (how much the precondition function error reduces with the refinement) and 2) Observation Difference Reduction Score ($ODRS$) (how much the action or dynamic function can reduce the observation difference after the refinement). We generate $K=3$ refinements for each error (using experiments with 50 train transitions, we notice refinements beyond 3 tend to be irrelevant, adding little value).

We use a validation set V (of 300 transitions randomly selected from the training data) to assess the quality of a refinement $r_t^j$ to the code $\phi'$ generated for a transition t in the training set.  We select the highest scoring refinement only if it shows an average improvement over the validation trajectories as well (we denote the function that averages $RefinementScore$ applied to validation trajectories as $ValidationScore$ ($VS$)). As mentioned in Algorithm \ref{alg:code_refinement}, we only choose refinements that have a $VS>0$ since any value less than 0 indicates that the refinement worsens the predictions of $\phi'$ for the validation set. 

Let an arbitrary transition in the validation set be ($s_{v-1}, x_{v-1}, \sigma_{v-1}, s_v, x_v$). Let $s_{v}'$, $\sigma_{v-1}'$ be the observation vector and action validity computed by $f_{\phi'}$ and $\hat{s_v}$ and $\hat{\sigma_{v-1}}$ be the observation vector and action validity computed after refinement $r_t^j$ is applied to $\phi'$ to get $\phi_{r_t^j}'$, i.e. $\hat{s_v}, \hat{\sigma_{v-1}} = f_{\phi_{r_t^j}'}(s_{v-1},x_{v-1}, \sigma_{v-1})$. Algorithm \ref{alg:refinement_score} explains the function's implementation in detail.

\section{Stochastic Component Implementation}\label{appendix:stochastic-implementation}
In this section we provide some additional implementation details for our stochastic component.
\begin{algorithm}[H]
\small
\caption{Sampling Semantically Coherent DAGs via LLM}
\label{alg:dag_sampling_semantic}
\begin{algorithmic}[1]
\Procedure{SemanticTopoSort}{$V_x, V_s, \mathcal{L}$}
    \State $\mathcal{O} \gets (v \text{ for } v \in V_s)$ \Comment{Initialize with exogenous variables}
    \State $V_{rem} \gets V_x$
    \While{$V_{rem} \neq \emptyset$}\Comment{While not all sorted}
        \State $P_1 \gets \text{ConstructPrompt}(\mathcal{O}, V_{rem}, \mathcal{L})$ 
        \State $v^* \gets \text{LLM}(P_1)$
        \State $\mathcal{O} \gets \mathcal{O} \oplus v^*$ \Comment{Append $v^*$ to topological order}
        \State $V_{rem} \gets V_{rem} \setminus \{v^*\}$
    \EndWhile
    \State \Return $\mathcal{O}$
\EndProcedure
\vspace{0.2cm}
\Procedure{ElicitDependencies}{$\mathcal{O}, V_x, V, \mathcal{L}$}
    \State $E \gets \emptyset$
    \For{$i \gets 1$ to $|\mathcal{O}|$}
        \State $v_i \gets \mathcal{O}[i]$
        \If{$v_i \in V_x$} \Comment{Skip vars in $V_s$}
            \State $C(v_i) \gets \{v_j \in V \mid v_j \text{ precedes } v_i \text{ in } \mathcal{O} \}$ 
            \State $P_2 \gets \text{ConstructPrompt}(\mathcal{L}[v_i], C(v_i), \mathcal{L})$
            \State $S \gets \text{LLM}(P_2)$ 
            \State $E \gets E \cup \{(v_p, v_i) \mid v_p \in S\}$\Comment{Directed edge}
        \EndIf
    \EndFor
    \State \Return $E$
\EndProcedure
\vspace{0.2cm}
\State \textbf{Input:} Set of nodes $V = V_x \cup V_s$, environment semantics $\mathcal{L}$
\State \textbf{Output:} A directed acyclic graph $G=(V, E)$
\vspace{0.1cm}
\State $\mathcal{O} \gets \text{SemanticTopoSort}(V_x, V_s, \mathcal{L})$
\State $E \gets \text{ElicitDependencies}(\mathcal{O}, V_x, V, \mathcal{L})$
\State \Return $G=(V, E)$
\end{algorithmic}
\end{algorithm}

\subsection{DAG Search}
\paragraph{Simulated Annealing:} We use a standard simulated annealing search to improve DAG structures. We generate candidate solutions by randomly considering graph modifications and only keeping ones that maintain acyclicity. We consider flipping a directed edge, removing one, and adding one. Alternatively, we could use other strategies to suggest actions, such as via word embedding similarity, which could improve the search.  We use a geometric cooling schedule with $\alpha=0.99$. We run 100 optimization steps per initial DAG for the coffeshop environment and 200 for MAPs.

\subsection{DAG Sampling}
\paragraph{Algorithm \ref{alg:dag_sampling_semantic}} shows the sampling procedure for DAGs encoding semantically coherent relationships between variables. First, a single pass over the variables is used to conduct a topological sort over the variables using the environment semantics. Given a partial topological sort, an LLM is instructed to select a variable that can be plausibly modeled using the variables sorted so far.  In the second step, another single pass over the variables uses an LLM to elicit variable dependencies, respecting the topological order. This ensures that the final edge set $E$ forms a DAG. While more complex procedures could be employed (e.g., by pairwise comparisons), we find that this approach works well in practice, especially when $\mathcal{L}$ is rich. Overall, this procedure requires $O(d)$ LLM calls, where $d$ is the dimension of $x_t$.

\paragraph{Implementation Details:} We use structured outputs and token log probabilities with a fixed temperature parameter to ensure consistent outputs when sampling DAGs from the LLM. Exact prompts are available in appendix \ref{appendix:prompts}. While more complex approximations of the prior using multiple LLM calls are possible, we use this single-prompt approach to improve efficiency, as this term is evaluated on every step of the search. For topological sorting, our structured format includes a field for thinking traces and a single string representing the next variable in the topological sort. For eliciting dependencies, our structured format contains the thinking field and a list of variable names chosen. Empirically, we notice that this tends to sample sparse graphs, as LLMs will tend to not answer with too many dependencies (likely due to context dilution when the number of variables is large). Still, we find that it is acceptable, as the search procedure in practice tends to add edges. 

\subsection{LLM prior} We use structured outputs to ensure consistent outputs from the LLMs. Exact prompts are available in appendix \ref{appendix:prompts}. Our structured output schema includes a field for thinking traces and a final answer as a Boolean variable. We extract log probabilities for the corresponding answer token and return it as the prior value. In practice, these values tend to be large (either for the token 'true' or 'false', depending on the answer provided). As such, the LLM provides a nearly hard constraint for the search, rejecting DAGs with semantically implausible dynamics. We use $\lambda_1=10$ to bring the sparsity factor in a similar scale to the pinball loss. We use $\lambda_2=100$. This turns the LLM prior into a steep constraint. The optimization first finds structures that match the environment dynamics ($\log p(E|w(V,\cal L) \approx 10^{-2}$). Then the search is concentrated among such graphs.

\section{Prompts}
\label{appendix:prompts}
\textbf{MAPs Environment Gameplay Rules:}
\begin{lstlisting}
## Game Mechanics

The purpose of the game is to maximize a theme park's profit given a starting budget and timeframe. As the CEO of the park, you perform one action at the start of the day. The park then opens and guests interact with the park for a full day, which consists of 500 ticks.

There are three difficulty modes to the game ("easy", "medium", and "hard").
  - Easy: All attractions are available from the beginning. Water tiles are disabled. Short time horizon (50 days by default)
  - Medium: Only yellow attractions are available from the beginning, other attractions must be researched. Layouts can include water tiles. Medium time horizon (100 by default)
  - Hard: Only yellow attractions are available from the beginning, other attractions must be researched. Layouts can include water tiles and terraforming actions (i.e. adding/removing paths/water) are enabled. Long time horizon (250 by default)

There are 7 primary components to the game. We provide a high level overview here, and a detailed description of each follows.
- The Park. This is defined by a square grid (defaults to 20x20) and contains all other components. The theme park has an entrance and an exit, which are connected by a path. 
- Terrain. There are three kinds of terrain: Paths, Water, and Empty. 
- Rides. Rides are one of two types of attractions you can place in your theme park. They are the core of the theme park, and are what draw guests in and keep them happy. There are three subtypes of rides: Carousels, Ferris Wheels, and Roller Coasters. Each has four subclasses: yellow, blue, green, and red.
- Shops. Shops are the second type of attractions you can place in your theme park. They allow you to cater your guests' needs. There are three subtypes of shops: Drink shops, Food shops, and Specialty shops. Drink shops alleviate the thirst of guests. Food shops alleviate the hunger of guests. Specialty shops provide a range of utilities. There similarly four subclasses for each subtype: yellow, blue, green, and red.
- Staff. Staff can be hired to maintain your park. There are two types of staff: janitors and mechanics. Janitors keep your park clean, while mechanics can repair rides that need maintenance. 
- Guests. This is who you build the park for! Guests enter your park to enjoy the attractions you've built.
- Research. Early on, you only know how to build certain rides. Investing in research will allow you to build more subclasses of rides and shops as time goes on.

A core feature of your park is your park rating. This rating depends on several factors in your park and is a driving force in how many guests will come visit your park.

### Terrain
- Empty tiles: A blank tile on which something can be built.
- Path tiles: The tiles used by guests to move around your park. All attractions must be placed on an empty tile that is adjacent to a path tile. 
- Water tiles: The excitement of any ride adjacent to a water tile is increased by 1, however it nothing can be built on it.

In hard mode, terrain can be terraformed. Paths can be built ($1000) and removed ($2500). Water tiles can similarly be built ($5000) and removed ($10000)

### Rides
Rides are the core of your theme park. They drive guest attraction and guest happiness. Rides have several key attributes:
- Capacity. How many guests can fit on your ride at a single time. The cummulative capacity of your park is also a key factor in how many guests can visit your park.
- Excitement. How exciting a ride is. Higher excitement scores lead to greater guest happiness. A high excitement score is also crucial to your park rating.
- Intensity. How intense the ride is. Keeping your average intensity balanced (i.e., as close to 5 as possible) will create a park that caters to a wide range of guests, and will help increase your park rating.
- Ticket price. How much money guests have to spend to ride the ride. You are here to make a profit after all. Note that if a guest does not have enough money to pay the ticket price, they will be rejected by the ride, which will decrease their happiness.
- Cost per operation. How much it costs each time you operate the ride.

Rides have no fixed costs, their operating costs depends solely on the number of times it is operated each day.  
Rides only operate if guests have boarded, and wait for a few turns after the first guest boards to see if more guests will join. This wait is longer for rides with a larger capacity, but decreases as more guests board the ride. A full ride will always operate.  
Rides will sometimes breakdown after operating. While broken down, it is out of service and will not accept guests. Having broken rides will negatively impact your park rating, so it is wise to hire mechanics to fix broken rides.  

There are three subtypes of rides: Carousels, Ferris Wheels, and Roller Coasters. Each subtype has distinct characteristics:
- Carousels a cheap to build, operate, and they rarely breakdown. However, they provide limited other benefits.
- Ferris Wheels are an intermediate ride option. They are the ride subtype with the highest capacities. 
- Roller Coasters are an expensive, but high-value ride. They boast the highest excitement and intensity scores, but breakdown more than the other types of rides.
    
Each ride also has four possible subclasses. In medium and hard mode, you only start with the yellow version of each ride, and have to perform research to unlock other subclasses of rides. Here is a general outline of subclasses
- Yellow rides are starter rides. Cheap to build and operate, but otherwise unremarkable.
- Blue rides provide an immediate step up. Often with higher excitement, intensity, capacity, and maximum allowable ticket prices.
- Green rides are the subsclass that provides the highest capacity, but have lower excitement and worse intensity values.
- Red rides are thrill rides. High excitement, high intensity, and have the highest possible ticket prices, but they prone to breaking down.

For the exact parameters of each ride, see [All Rides](#all-rides)

### Shops
Shops are necessary to adequately cater to guest needs. Shops have several key attributes:
- Operating costs: how much it cost to stock the shop each day. Unlike rides, shops cost a fixed amount per day.
- Item price: the price guests pay for the item being sold.

Similar to rides, there are three subtypes of shops:
- Drink. Drink shops sell drinks that quench the thirst of guests.
- Food. Food shops sell food that satiate the hunger of guests.
- Specialty. Specialty shops provide a range of services based on their subclass. See [All Shops](#all-shops) for a description of each. Notably, guests will never seek out specialty shops, they will only visit specialty shops if they walk by them.

Again similar to rides, shops have subclasses:
- Yellow food and drink shops are starter shops. They provide a basic version of their product. 
- Blue food and drink shops sell an improved (more hunger satiating / thirst quenching) version of their respective product.
- Green food and drink shops provide multiple benefits instead of a single benefit.
- Red drink shops sell coffee which caffeinate guests. Red food shops are luxury food items, providing highly satiating food that also boost happiness.

For exact parameters of each shops, see [All Shops](#all-shops)

### Staff
There are two types of staff: a janitor and a mechanic. Each staff has a salary: $25 for mechanics and $100 for janitors. 

Janitors will roam the park, generally moving toward dirtier areas. When on dirty tiles, janitors will clean them.

Mechanics will move toward rides that are broken down, with a preference towards nearer broken rides. When they reach a tile containing a ride that needs maintenance, mechanics will perform repairs until the ride is operational again. Rides will require 0.05%

Multiple staff can occupy the same tile. Multiple of the same kind of staff on a tile will multiplicatively increase the speed at which the tile is cleaned / repaired.

### Guests
Guests are what the park is made for. The amount of guests that visit your theme park is based on the park's rating and the overall capacity of the park.
*Capacity* determines both how many guests can be in the park at any given time, as well as how many potential guests consider visiting the park.
Capacity is entirely determined by the cummulative capacity of the current rides in the park. 
**NOTE:** Since only rides increase capacity, a park with no rides will receive no visitors. We aren't making a food hall.

*Park Rating* determines the likelihood that a potential guest decides to enter the park and become a real guest. New guests cannot enter the park if the park is currently at capacity.
Park rating can be increased by increasing the total excitement of the park, and by having guests leave your park happy. Park rating will drop if the park is dirty, rides have low uptimes (i.e., are out of service for portions of the day), or if the average intensity of rides is too high or too low.

Each guests that visits your park will bring with them some amount of money. If they run out of money, they will leave your park.
Each guests also has a finite amount of energy. Every step they take in the park will decrease this energy. If they run out of energy, they will leave your park.

Guests also have hunger, thirst, and happiness levels. Hungry guests will seek out food shops. Thirsty guests will seek out drink shops. Unhappy guests will seek out rides. If a guest's hunger or thirst levels get too high, it will negatively impact their happiness. If guests become too unhappy, too thirsty, or too hungry, they will leave your park. 

If a guest has all their needs met, they will target any attraction (except specialty shops, see below). If guests pass by another attraction on their way to their target, they may visit that attraction. Guests like novelty and will favor attractions they have visited less frequently. Guests will also favor attractions that are nearby, but they will never visit the same attraction twice in a row. If a guest visits an attraction that they cannot afford or that is currently broken down, their happiness will be affected.
**NOTE:** Guests will never seek out specialty shops. They will only visit specialty shops if they pass by them on their way to another attraction.

Guests sometimes litter. This decreases the cleanliness of the path or attraction they are currently on. Unhappy guests are more likely to litter. If guests visit dirty tiles, it negatively impacts their happiness. If a guest visits a ride that is too dirty, they may choose to go elsewhere which will negatively impact their happiness.

In hard mode, guests may have preferences, meaning they will only interact with a subset of attractions. If a guests visits a ride that does not match their preference, it will negatively impact their happiness. Providing guests with information through an information booth (blue specialty shop) will allow them to know ahead of time which attractions match their preference.

**NOTE:** You can learn more about guests by surveying them through the SurveyGuest action. This provides information about why the guest left the park, what the guest preferences are, and more. You can choose how many guests to survey (up to a maximum of 25) for a price of $1000 per guest.

### Research
In easy mode, all attractions are avaiable from the start and research is disabled. This section is only relevant to medium and hard mode.

At the start of the game, you only the yellow subclass of each attraction is available. To learn how to build more subclasses, you have to invest in research. To do this, you have to set your research, which includes setting how fast you want to perform research and the topic(s) of research (where the topics are attraction subtypes). Once set, research will be performed each day according to your settings. This persists until you change the research settings, until you run out of funds, or you successfully research all possible subclasses for the specified research topics. In the latter two cases, the research speed will be set back to "none". Research will always unlock new subclasses in the following order: blue, green, red. Once a new attraction has been unlocked, research will continue on the next topic in your list of topics.

**NOTE:** if, for example, you want to unlock the red roller coaster as fast as possible, you will want to set the research speed to "fast" and set your research topics to ONLY ["roller coaster"].

Performing research faster requires more money. Below are the research speeds and their rates:
- "none": 0$/day; research is halted at this speed.
- "slow": 2000$/day.
- "medium": 10000$/day.
- "fast": 50000$/day

The default setting is a research speed of "none" and all attraction subtypes selected as topics. 

## All Rides

### Carousels

**Yellow**
- Building Cost: 250
- Cost per Operation: 1
- Capacity: 7
- Max Ticket Price: 3
- Excitement: 1
- Intensity: 1
- Breakdown Rate: 0.001

**Blue**
- Building Cost: 1250
- Cost per Operation: 2
- Capacity: 14
- Max Ticket Price: 6
- Excitement: 4
- Intensity: 5
- Breakdown Rate: 0.002

**Green**
- Building Cost: 7500
- Cost per Operation: 16
- Capacity: 26
- Max Ticket Price: 7
- Excitement: 2
- Intensity: 4
- Breakdown Rate: 0.003

**Red**
- Building Cost: 15000
- Cost per Operation: 12
- Capacity: 18
- Max Ticket Price: 11
- Excitement: 7
- Intensity: 5
- Breakdown Rate: 0.005

### Ferris Wheels

**Yellow**
- Building Cost: 500
- Cost per Operation: 6
- Capacity: 10
- Max Ticket Price: 4
- Excitement: 3
- Intensity: 2
- Breakdown Rate: 0.006

**Blue**
- Building Cost: 3750
- Cost per Operation: 8
- Capacity: 22
- Max Ticket Price: 5
- Excitement: 6
- Intensity: 3
- Breakdown Rate: 0.009

**Green**
- Building Cost: 37500
- Cost per Operation: 40
- Capacity: 42
- Max Ticket Price: 9
- Excitement: 5
- Intensity: 7
- Breakdown Rate: 0.023

**Red**
- Building Cost: 55000
- Cost per Operation: 28
- Capacity: 24
- Max Ticket Price: 12
- Excitement: 9
- Intensity: 8
- Breakdown Rate: 0.018

### Roller Coasters

**Yellow**
- Building Cost: 1000
- Cost per Operation: 10
- Capacity: 5
- Max Ticket Price: 15
- Excitement: 4
- Intensity: 6
- Breakdown Rate: 0.01

**Blue**
- Building Cost: 10000
- Cost per Operation: 25
- Capacity: 8
- Max Ticket Price: 20
- Excitement: 8
- Intensity: 8
- Breakdown Rate: 0.02

**Green**
- Building Cost: 30000
- Cost per Operation: 40
- Capacity: 16
- Max Ticket Price: 18
- Excitement: 6
- Intensity: 9
- Breakdown Rate: 0.03

**Red**
- Building Cost: 75000
- Cost per Operation: 50
- Capacity: 10
- Max Ticket Price: 50
- Excitement: 10
- Intensity: 10
- Breakdown Rate: 0.033

## All Shops

### Drink Shops

**Yellow**
- Building Cost: 100
- Operating Cost: 16
- Max Item Price: 2
- Thirst Reduction: 0.5

**Blue**
- Building Cost: 1750
- Operating Cost: 75
- Max Item Price: 6
- Thirst Reduction: 0.9

**Green**
- Building Cost: 17500
- Operating Cost: 280
- Max Item Price: 10
- Thirst Reduction: 0.8
- Happiness Boost: 0.4

*In addition to quenching thirst, green drink shops additionally provide a boost to guest happiness.*

**Red**
- Building Cost: 48000
- Operating Cost: 600
- Max Item Price: 15
- Thirst Reduction: 0.4
- Energy Boost: 50
- Caffeinated Steps: 50

*Red drinks caffeinate guests, which boosts a guest's energy and allows them to move twice as fast*

### Food Shops

**Yellow**
- Building Cost: 150
- Operating Cost: 28
- Max Item Price: 4
- Hunger Reduction: 0.4

**Blue**
- Building Cost: 3000
- Operating Cost: 150
- Max Item Price: 10
- Hunger Reduction: 0.8

**Green**
- Building Cost: 32000
- Operating Cost: 500
- Max Item Price: 18
- Hunger Reduction: 0.6
- Thirst Reduction: 0.6

*Green food shops both satiate hunger and quench thirst.*

**Red**
- Building Cost: 60000
- Operating Cost: 1000
- Max Item Price: 25
- Hunger Reduction: 0.9
- Happiness Boost: 0.5

*Red food shops sell luxury food. This greatly satiates hunger and increases happiness*

### Specialty Shops

NOTE: Guests will not target specialty shops, and will only visit specialty shops if the walk adjacent to one.

**Yellow (Souvenir Shop)**
- Building Cost: 250
- Operating Cost: 80
- Max Item Price: 12
- Happiness Boost: 0.3

*These provide a happiness boost to guests that buy them the first time, but this happiness boost diminishes with each subsequent souvenir purchased.*

**Blue (Information Booth)**
- Building Cost: 10000
- Operating Cost: 200
- Max Item Price: 5

*These provide information to guests about the rides in the park and ensure that guests only visit rides that fall within their budget and preferences. These do not provide guests about information related to the cleanliness or operation (i.e., if an attraction is out of service) of an attraction.*

**Green (ATM)**
- Building Cost: 50000
- Operating Cost: 500
- Max Item Price: 2
- Money Withdrawal: 50

*ATMs allow guests to withdraw more money. The amount of money withdrawn decreases exponentially with every subsequent withdrawal, to a minimum of 5*

**Red (Billboard)**
- Building Cost: 10000
- Operating Cost: 0
- Max Item Price: 0
- Thirst Boost: 0.5
- Hunger Boost: 0.5

*Billboards make guests more hungry and thirsty, will reset the visit count of attractions (meaning that guests are more likely to revisit attractions), and, if the guest has less than $20, will set the guest's target to an ATM.*
\end{lstlisting}

\textbf{CoffeeShopSimulator Environment Gameplay Rules:}
\begin{lstlisting}
    """
Welcome to the manager's seat! Running a successful coffee shop is an art. Here's what you need to know to become a local legend.

The Heart of the Business: Customer Happiness

Your central goal is keeping customers happy. A shop's **satisfaction** level is the key to everything. This mood is shaped by the **quality of your shop**upgrades make a big difference and the **price** you set. Finding the right balance is crucial, as customers have strong opinions about what a coffee should cost. Remember, your reputation lingers, so yesterday's vibe will influence today's crowd.

Drawing a Crowd

The number of customers you attract is a direct result of their happiness. A buzzing, satisfied shop creates a **word-of-mouth** effect that tends to increase your future customers! Your shop's **upgrade level** also determines its basic appeal and how many potential customers are in the area. However, be prepared! You can only serve as many people as you have **beans and milk** for, so a sudden rush can leave you empty-handed if you don't plan ahead.

The Daily Ledger

At the end of each day, you'll see the results of your hard work. The number of customers you served and the price you charged will determine your **revenue**, but don't forget you have daily costs to cover. Every cup sold uses up your precious **inventory**, and every visitor makes the shop a little less **clean**. Mastering this daily cycle of cause and effect is your path to building a coffee empire.

"""
\end{lstlisting}
\subsection{Code Optimization}\label{appendix:code_prompts}
\textbf{Environment Background Info - TPT}
\begin{lstlisting}
*** Environment Background ***
You are working with a theme park world model. The infomation about the state of the world is represented in JSON format below.
You are also given a set of actions that can be taken to change the state of the world.
Finally, you are given a description of the Theme Park environment that is modeled.
                        
***Park State Specification***

The theme park is represented by {PARK_SIZE} x {PARK_SIZE} cartesian grid.
The state of theme park is represented in the following json structure.

State Specification:
{STATE_SPEC}

Action Specification:
{ACTION_SPEC}

***Gameplay Rules***
{GAMEPLAY_RULES}

Below is an example of how a state of the world is represented:
{EXAMPLE_STATE}

Below are examples of how some actions may initialize objects in the state.
{INITIALIZERS}
\end{lstlisting}

\textbf{JSON Patch Instructions}
\begin{lstlisting}
### How to Use JSON Patch to Define Changes to JSON Files

**JSON Patch** (based on [RFC 6902] is a standard format for describing updates to a JSON document. A **JSON Patch** document is an array of operations. Each operation tells you what to change and where.

---

## 1. Structure of a JSON Patch

Each patch operation is a JSON object with:

| Field   | Description                               
| `op`    | Operation to perform: `add`, `remove`, `replace`, `move`, `copy`, or `test` |
| `path`  | JSON Pointer (`/path/to/key`) to the location in the document. Array indices are specified as `/path/to/key/0`. If the array is empty, use index 0.                  |
| `value` | (Optional) New value for `add`, `replace`, `test`                           |
| `from`  | (Required for `move` and `copy`) The source location                        |

---

## 2. Supported Operations (with Examples)

Let's say you have a state that has a list of cars in the path /car_info/cars. We can work with this example to understand the operations. 
Note that the path is a JSON Pointer, so you need to use the correct syntax for the path.

### `add`

Adds a value to an object or array.

When a variable that should be an array is empty, create it first.

Example: Adding a new car to the list of cars.
```json
{
  "op": "add",
  "path": "/car_info/cars",
  "value": [
    {
      "id": "car-1",
      "make": "Toyota",
      "model": "Camry",
      "year": 2020
    }
  ]
}
```

To add a new member to the cars variable, below is an example:
```json
{
  "op": "add",
  "path": "/car_info/cars/2",
  "value": {
    "id": "car-2",
    "make": "Toyota",
    "model": "Camry",
    "year": 2020
  }
}
```
Never simply add an empty dictionary to the list. Always add a fully populated dictionary.
If the path exists, it replaces the current value (like `replace`). If it doesn't, it adds the new value.
**Note that when adding to an array, the index must be specified. Do not overwrite the entire array.**
**Empty arrays don't have any index. If there are no cars, the path is `/car_info/cars`.**

---

### `remove`

Deletes a value from the document.

```json
{
  "op": "remove",
  "path": "/car_info/cars/0"
}
```

Removes the first car (index 0).

---

### `replace`

Replaces the value at the given path. Only replace INDIVIDUAL VALUES. Do not replace the entire dict object!

```json
{
  "op": "replace",
  "path": "/car_info/cars/0/make",
  "value": "Ford"
}
```

Changes the `make` field to "Ford".

---

## 3. Path Syntax (JSON Pointer)

* `/foo' $\rightarrow$ the key `"foo"` at the root level
* `/foo/0` $\rightarrow$ the first element in the array under key `"foo"`
* `/a~1b` $\rightarrow$ refers to key `a/b` (because `/` is escaped as `~1`)
* `/m~0n` $\rightarrow$ refers to key `m~n` (because `~` is escaped as `~0`)

---
Make sure you produce a valid JSON Patch. This includes following the JSON Patch specification, and expressing it with a valid json. Do not include any raw computations in your prediction. Do not include comments or other text in your prediction as it will be parsed automatically.
Return all the operations in a single JSON Patch array.
\end{lstlisting}

\textbf{Action Function Format}:
\begin{lstlisting}[language=]
*** Action Function Format ***
The action function is a Python function that takes the current state and action as input, and returns a JSON patch that captures the changes that the action makes to the state.
The format for a JSON patch is described in the instructions below:
{patch_instructions}

*Guidelines:*
  - The function should be implemented to strictly follow the description of the function provided.
  - Make sure the function sticks to the state and action specifications provided.
  - Don't use any libraries other than the standard Python libraries.

**Output Format**
First plan the function implementation step by step within <thought> and </thought> tags accounting for all the edge cases.
Then generate the code for the function in <code> and </code> tags.

<code>
```python
def action_name(state, action): 
    # Your code here
    return <JSON patch dict> 
```
</code>
\end{lstlisting}

\textbf{Precondition Function Format}:
\begin{lstlisting}[language=]
*** Precondition Function Format ***
The precondition function is a Python function that takes the current state and action as input, and returns a boolean value indicating whether the action can be executed in the current state.
It also returns a feedback string that explains why the action cannot be executed if it returns False. Feedback simply says "Precondition does not fail" if the action can be executed.

*Guidelines:*
  - The function should be implemented to strictly follow the description of the function provided.
  - Make sure the function sticks to the state and action specifications provided.
  - Don't use any libraries other than the standard Python libraries.

**Output Format**
First plan the function implementation step by step within <thought> and </thought> tags accounting for all the edge cases.
Then generate the code for the function in <code> and </code> tags.

<code>
```python
def precondition_function_name(state, action): 
    # Your code here
    return <boolean value>, <feedback string>
```
</code>
\end{lstlisting}

\textbf{Dynamic Function Format}:
\begin{lstlisting}[language=]
*** Dynamic Action Format ***
The dynamic action is a Python function that takes the current state as an input and returns a JSON patch that captures the change to the state due to some continuous event (independent of any action).
The format for a JSON patch is described in the instructions below:
{patch_instructions}
*Guidelines:*
  - The function should be implemented to strictly follow the description of the function provided.
  - Make sure the function sticks to the state and action specifications provided.
  - Don't use any libraries other than the standard Python libraries.

**Output Format**
First plan the function implementation step by step within <thought> and </thought> tags accounting for all the edge cases.
Then generate the code for the function in <code> and </code> tags.

<code>
```python
def dynamic_function_name(state): 
    # Your code here
    return <JSON patch dict> 
```
</code>
\end{lstlisting}

\textbf{Code Initialization Prompt Template ($T_{init}$)}:\\
\textbf{Generate Action and Precondition Functions}\\
\textit{System Prompt Template}
\begin{lstlisting}[language=]
You are a world model agent for a given environment. The background of the environment is provided below.
{background_info}.
You need to generate a complete action function implementation including both the description and the Python code. An action function is a Python function that takes the current state and action as input, and returns a JSON patch that captures the changes that the action makes to the state.
You also need to generate complete precondition function implementations for each action. A precondition function is a Python function that takes the current state and action as input, and returns a boolean value along with a string feedback indicating whether the action can be executed in the current state.
The functions should only change the state varibles in the deterministic state specification provided below:
{DETERMINISTIC_STATE_SPEC}
A function description is a JSON with the following keys:
name: <function_name>
purpose: <function_purpose>
implementation_details: <function_implementation_details>

The format for action function is described below:
{action_function_format}
The format for precondition function is described below:
{precondition_function_format}

You need to generate a complete action function implementation for the action specification provided.
Note: The precondition functions will check for the preconditions, action function can assume that the preconditions are SATISFIED.
Remember to keep the action description as general as possible, so that it can be used in any state of the world model.

Think carefully and write all preconditions that need to be satisfied. Remember to keep the preconditions as general as possible, so that they can be used in any state of the world model.
The preconditions should be granular, each checking for a single aspect of the state change.

Note: Changes to the state independent of any action are captured by a dynamic function that is executed after the action function. So, don't include any logic in the action function that is not an affect of the action.

**Output Format**
Return a JSON with the following keys and values:
"action_name": <action_name>,
"action_description": <action_function_description>,
"code": <action_function_code>,
"preconditions": [<precondition_1_output>, <precondition_2_output>, ...]

Where each precondition output contains:
"precondition_description": <precondition_function_description>,
"code": <precondition_function_code>
\end{lstlisting}
\textit{User Prompt Template}
\begin{lstlisting}[language=]
Now generate the complete action function implementation and precondition function implementations for the following action:
{action_spec}
\end{lstlisting}

\textbf{Generate Dynamic Function}
\textit{System Prompt Template}
\begin{lstlisting}[language=]
You are a world model agent for a given environment. The background of the environment is provided below.
{background_info}.
You need to generate a complete dynamic function implementation including both the description and the Python code. 
The world model agent needs to be able to predict the next state given the current state. This can only be done by capturing default changes that happen in the environment dict which is not influenced by any action being taken.

A function description is a JSON with the following keys:
name: <function_name>
purpose: <function_purpose>
implementation_details: <function_implementation_details>
The format for dynamic function is described below:
{dynamic_function_format}
When generating the dynamic function, only consider computing changes to the elements of the deterministic state specification provided below:
{DETERMINISTIC_STATE_SPEC}
Note: An action function may be applied to the state first, then the patch generated by the dynamic function is applied to the modified state to compute the next state's deterministic values correctly.

Only include logic in the dynamic function if it is needed to compute the correct next state.
Note: The dynamic function should capture all the continuous events that happen in the world model. If multiple continuous events are happening at the same time, combine them into a single dynamic function.

**Output Format**
Return a JSON with the following keys and values:
"dynamic_description": <dynamic_function_description>,
"code": <dynamic_function_code>
\end{lstlisting}
\textit{User Prompt Template}
\begin{lstlisting}[language=]
Now look at the deterministic state specification and generate a complete dynamic function implementation.
\end{lstlisting}

\textbf{Code Refinement Prompt Template ($T_{refine}$)}\\
\textit{System Prompt Template:}
\begin{lstlisting}[language=]
You are a World Model Agent for a given environment. The background of the environment is provided below.
{background_info}.
You are using Python functions to predict the changes to the world state based on an action taken, which is called a transition.
You are given one of more functions that are currently being used and need to be refined. Refinement is done to address shortcomings in the functions which will be described.
A refinement can have the following forms: Adding a new function, removing an existing function, or replacing an existing function.
You are also given context. Take your time to understand the context and identify the shortcomings with the function set before proposing the refinements.
You are also given the number of refinements you need to propose. Think carefully and return the best refinements you can come up with.

Refinements can have the following forms: Adding a new function, removing an existing function, or replacing an existing function.

Assume that ALL the important environment details required to predict the changes to the world state are provided in the background.

Functions might need to produce JSON patches to update the state. The format for a JSON patch is described in the instructions below:
{patch_instructions}

While the functions can use all the variables in the state spec, they only need to correctly compute the changes to the variables in deterministic values (in state spec given below).
{DETERMINISTIC_STATE_SPEC}

The formats of the three types of Python functions are provided below (stick to this format depending on the type of function you are adding/replacing):
{action_function_format}
{precondition_function_format}
{dynamic_function_format}

When generating code, make sure to include both the function description and the complete Python code implementation (that handles necessary edge cases).

Note: Here is how the functions are applied to the state:
- First, preconditions are checked. If they are satisfied, action function is executed and the resulting patch is applied to the state.
- Then, dynamic function is executed and the resulting patch is applied to the modified state. (Even if the action function patch is not applied, the dynamic function patch is still applied.)

Precondition functions can be granular, focusing on a narrow part of the state change. But the action function should cover all changes an action makes to the state, and the dynamic function should cover all continuous changes that happen in the world model.

Make sure to use the correct 'function_id' for the function to be removed or replaced. Stick to the format expected for the output.
**Output Format**
Return a JSON of refinements.
\end{lstlisting}

\textit{User Prompt Template}
\begin{lstlisting}[language=]
The required information is provided below:
The type of error is: {error_type}
The context:
{context}
Context functions:
{functions}

Now generate the most useful refinements for the function(s) given:
\end{lstlisting}

\subsection{Graphical learning}\label{appendix:stochastic-prompts}

\textbf{Topological Sort Prompt}\\
\textit{System Prompt Template:}
\begin{lstlisting}[language=]
You are an expert at the {name} game. Here is the game manual:

{game_manual}

Your task is to determine, based on the game manual and your knowledge of the game, how game variables relate to each other in the current state.
\end{lstlisting}

\textit{User Prompt Template:}
\begin{lstlisting}[language=]
In the game {game name}, which variables at time t would you use to compute the value of {variable_name}: {variable_description} at the same time t, given values of {temporal_parents} at time t-1 and the selected action. Candidate variables at time t: {candidate_variables}
\end{lstlisting}

\textbf{Causally Plausibe Relationship Prompt Template ($w(\mathcal{L}, V, E)$)}\\
\textit{System Prompt Template:}
\begin{lstlisting}[language=]
You are an expert machine learning engineer and theme-park game player. Here is the game manual:

{gameManual}. 

Base your responses on the game manual as well as your general knowledge of the world.
\end{lstlisting}

\textit{User Prompt Template:}
\begin{lstlisting}[language=]
I need help forecasting the variable {targetVariable_name}: {targetVariable_description} in game {name} using a graphical model. I need your help deciding whether the design of the graphical model dependencies are causally valid based on the game manual and your knowledge of the game. It is okay if some of the dependencies are not mentioned in the manual, as long as they are sensible. To predict {targetVariable_name} at step t, i use the values of {temporal_parents_str} at t-1. This is the list of candidate variables: {all_variables_str}. Answer True if the dependencies are plausible and False otherwise. Note that all dependencies need to be at least plausible. If any dependencies are not plausible, then say False. Always think step by step, considering each variable one at a time. 
\end{lstlisting}

\end{document}